\def\year{2022}\relax
\documentclass[letterpaper]{article} 
\usepackage{aaai22}  
\usepackage{times}  
\usepackage{helvet}  
\usepackage{courier}  
\usepackage[hyphens]{url}  
\usepackage{graphicx} 
\usepackage{natbib}  
\usepackage{caption} 
\DeclareCaptionStyle{ruled}{labelfont=normalfont,labelsep=colon,strut=off} 
\usepackage{algorithm}
\usepackage{algorithmic}

\usepackage{bm}
\usepackage{amsfonts}
\usepackage{amsmath}
\usepackage{mathtools}
\usepackage{array}

\usepackage{url,subfigure,amsmath,amssymb,epsfig,verbatim,booktabs,graphicx,epstopdf}
\usepackage[colorlinks=false,linkcolor=black, citecolor=blue, urlcolor=black, pdfborder={0 1 0}]{hyperref}
\usepackage{mathtools}
\usepackage{threeparttable}
\usepackage{multirow}
\usepackage{makecell}

\usepackage{color}
\usepackage{xcolor}

\usepackage{newfloat}
\usepackage{listings}
\floatstyle{ruled}
\newfloat{listing}{tb}{lst}{}
\floatname{listing}{Listing}
\title{Learning to Solve Routing Problems via Distributionally Robust Optimization}
\author{
    Yuan Jiang,\textsuperscript{\rm 1\equalcontrib}
    Yaoxin Wu,\textsuperscript{\rm 1\equalcontrib}
    Zhiguang Cao,\textsuperscript{\rm 2\thanks{Zhiguang Cao is the corresponding author.}}
    Jie Zhang\textsuperscript{\rm 3}
}
\affiliations{
    \textsuperscript{\rm 1}SCALE@NTU  Corp  Lab, Nanyang  Technological University, Singapore\\

    \textsuperscript{\rm 2}Singapore Institute of Manufacturing Technology, A*STAR, Singapore\\
    \textsuperscript{\rm 3}School of Computer Science and Engineering, Nanyang Technological University, Singapore\\
    jiang.yuan@ntu.edu.sg, wuyaoxin@ntu.edu.sg, zhiguangcao@outlook.com, zhangj@ntu.edu.sg
}

\begin{document}

\maketitle

\begin{abstract}
\begin{quote}
Recent deep models for solving routing problems always assume a single distribution of nodes for training, which severely impairs their cross-distribution generalization ability. In this paper, we exploit group distributionally robust optimization (group DRO) to tackle this issue, where we jointly optimize the weights for different groups of distributions and the parameters for the deep model in an interleaved manner during training. We also design a module based on convolutional neural network, which allows the deep model to learn more informative latent pattern among the nodes. We evaluate the proposed approach on two types of well-known deep models including GCN and POMO. The experimental results on the randomly synthesized instances and the ones from two benchmark dataset (i.e., TSPLib and CVRPLib) demonstrate that our approach could significantly improve the cross-distribution generalization performance over the original models.
\end{quote}
\end{abstract}

\section{Introduction}
Combinatorial optimization problems (COPs) with NP-hardness are always featured by discrete search space and intractable computation to seek the optimal solution. As a fundamental COP in logistics, the vehicle routing problem (VRP)~\cite{dantzig1959truck} concerns the cost-optimal delivery of items from the depot to a set of geographically scattered customers through vehicle(s). It has been extensively investigated for decades and found wide-spread applications in reality, such as waste collection~\cite{han2015waste}, dial-a-ride~\cite{malheiros2021hybrid} and courier delivery~\cite{steever2019dynamic}. 

The studies on VRP with deep (reinforcement) learning is emerging in recent years. Different from the conventional methods, this line of works aims at automatically searching heuristic policies by using neural networks to learn the underlying patterns in instances, which could be used to discover better policies than hand-crafted ones~\cite{bengio2021machine}.
Towards reducing the gaps to the highly optimized conventional heuristic solvers including Concorde~\cite{applegate2006concorde} and LKH~\cite{helsgaun2000effective}, a large number of efforts have been performed to invent various deep models to solve the VRP variants, i.e., traveling salesman problem (TSP) and capacitated vehicle routing problem (CVRP)~\cite{khalil2017learning,kool2018attention,chen2019learning,hottung2019neural,ma2021learning,wu2021learning,kwon2020pomo,li2021hcvrp,xin2021neurolkh}.



Although much success has been achieved, existing deep models always assume a pure spatial distribution of nodes (customers) for training, i.e., uniform distribution, which severely limits their applications given the impaired cross-distribution generalization ability. While it is natural to stipulate that a majority of typical instances follow an identical distribution, a minority of atypical (yet important) instances which follow different one(s) may always exist in reality~\cite{hashimoto2018fairness}. In this sense, the mono-training paradigm in existing deep models may cause inferior performance or even failures for those atypical instances. E.g., a trained routing policy may offer unreasonable routes to serve a group of (important) customers whose location pattern differs from the ones used in training.  Intuitively, an alternative is to force the model to homogeneously treat all the instances of different distributions for training. However, it does not necessarily enhance the cross-distribution generalization performance in our view, as it may suffer high losses on the group of the atypical instances. 




To address this issue, we propose an approach by exploiting \emph{group distributionally robust optimization} (group DRO) to jointly train the deep models on instances of different groups (more than one group), where each group follows a distribution. In particular, we aim at minimizing the loss for the \emph{worst-case} group during training, where we optimize the weights for different groups of instances and the parameters for the deep model in an interleaved manner. More importantly, we do not need to label the distribution for each group during inference. In addition, we also leverage convolution neural network (CNN) to learn initial representations of VRP instances so that the distribution-aware features in spatial patterns could be favorably captured to boost the performance further. The experimental results show that, our approach not only achieves superior performances on the overall instances and the worst-case performance on the atypical instances, but also exhibits desirable applicability to different deep models. Note that we do not claim to surpass the state-of-the-art deep methods for solving VRPs in all aspects, but to inject them with robustness for better cross-distribution generalization instead. Our contributions in this paper are summarized as follows.

\begin{itemize}
    \item We exploit group DRO to add the dimension of robustness to deep models, which enhances their cross-distribution generalization for solving VRPs. The proposed approach could be freely applied to various deep models, in the fashions of either supervised or reinforcement learning, e.g., GCN~\cite{joshi2019efficient} and POMO~\cite{kwon2020pomo}, respectively.
    
    \item We leverage CNN to initially identify the spatial pattern of the nodes in VRP instances, which allows the deep models to learn more informative distribution-aware representation and thus generate solutions of higher quality.
    
    \item We apply our approach to solve randomly generated instances of different distributions. The results show that it not only improves the overall performance and the worst-case performance over the original deep models, but also achieves superior performance for solving the instances from the benchmark dataset, i.e., TSPLib and CVRPLib.
    
\end{itemize}

\section{Related Works}
In this section, we provide a review of the deep models for solving VRPs, and the DRO in the machine learning community, respectively. 

\subsection{Deep Models for VRPs} 

The recent learning-based methods have shown great merits to automatically discover heuristics for solving VRPs, which effectively circumvent massive human expertise and trial-and-error required in the classic hand-engineering methods. Most of the deep models concentrate on solving TSP and CVRP, i.e., two representative VRP variants. The very early endeavor was the Pointer Network which used a recurrent neural network (RNN) to learn node selection in a supervised manner \cite{vinyals2015pointer}. Similarly, \citeauthor{joshi2019efficient} (\citeyear{joshi2019efficient}) predicted edges which will appear in the optimal solutions with supervised learning, where a graph convolutional network (GCN) was developed for deep embedding of both nodes and edges. The other works diverged mainly by employing different Transformer-based architectures and training with reinforcement learning \cite{Bello2017WorkshopT,kool2018attention,xin2021multi}. Specially, \citeauthor{khalil2017learning} (\citeyear{khalil2017learning}) tackled TSP using a graph embedding network. Instead of learning heuristics to select nodes, another line of works sought policies to improve solutions with local search frameworks~\cite{chen2019learning,lu2019learning,hottung2019neural,wu2021learning}. Among the deep models, \citeauthor{kwon2020pomo}  (\citeyear{kwon2020pomo}) presented a method called POMO to train multiple rollouts on augmented instances and delivered the state-of-the-art performance. On the other hand, despite the near-optimal results, most of the above deep models are validated with the synthesized instances of the uniform distribution. The cross-distribution generalization is not explicitly considered in their training process, which may degenerate on non-uniform instances, e.g., the ones from the well-known TSPLib and CVRPLib. 

\subsection{Distributionally Robust Optimization}

It is known that the standard maximum likelihood estimation may degrade the inference performance of deep models, if the training samples are out of the distribution for test~\cite{oren2019distributionally,kuhn2019wasserstein}. This issue about the out-of-distribution generalization is still stubborn for the general learning models, either in supervised or reinforcement learning~\cite{sun2019test,cobbe2019quantifying}.

One of the solutions to ameliorate the generalization capability is using distributionally robust optimization (DRO), which seeks to minimize losses over all sub-populations of the training distribution~\cite{ben2013robust,duchi2019distributionally}.  It was originally designed for classic underparameterized models to reduce the training loss, by regularizing the models and defending them against adversarial examples~\cite{namkoong2017variance,sinha2017certifiable}. Different from them, group DRO instead defines the uncertainty set as mixtures (or combinations) of groups over training data to avoid optimizing implausible worst-case groups~\cite{hu2018does,oren2019distributionally}. In this line, \citeauthor{sagawa2019distributionally} (\citeyear{sagawa2019distributionally}) applied group DRO in the overparameterized regime with vanishing training loss and poor worst-case generalization, and suggested that desirable accuracy for the worst-case group could be attained even if the groups are imperfectly designated. In this paper, we exploit group DRO to enhance the cross-distribution generalization of deep models for solving VRPs, where we explicitly define the uncertainty via (instance) groups of different distributions and minimize the worst expected loss over them.

\begin{figure*}
  \centering
  \includegraphics[width=\linewidth]{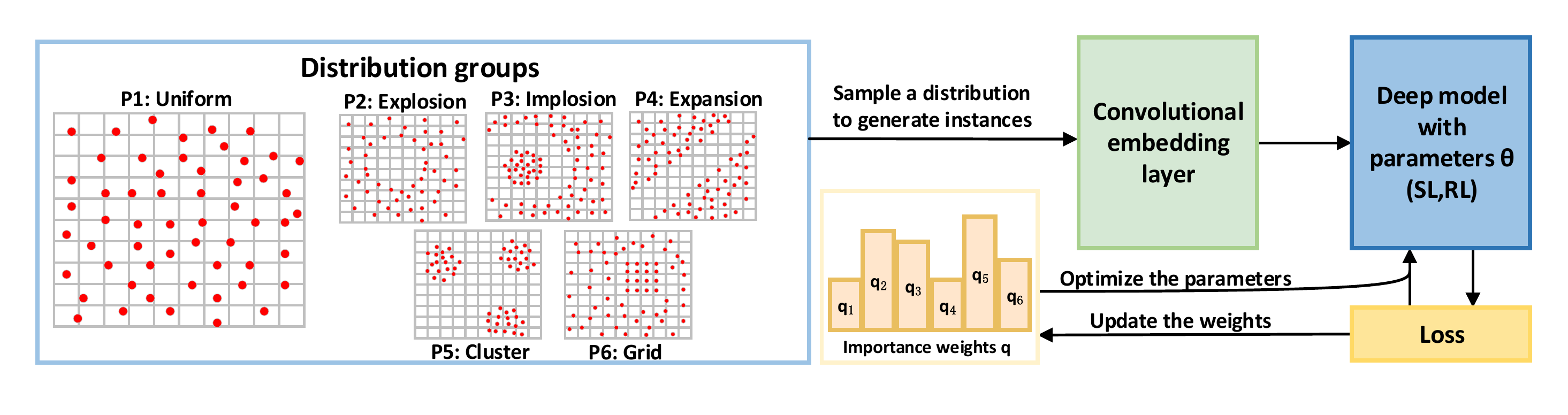}
  \caption{An example illustration of our approach with VRP instances sampled from six distribution groups, which alternately optimizes the parameters $\theta$ of a deep model and updates the importance weights $q$ of the distribution groups in the training set.}
  \label{fig:procedure}
\end{figure*}

\section{Preliminaries}

In this section, we present the graph representation of TSP and CVRP, followed by the basic rationale of DRO.

\subsection{VRPs and Data}
We consider TSP and CVRP in 2D Euclidean space. A TSP instance is represented as a fully connected graph $G$ with $n$ nodes $\left\{v_{1}, v_{2}, \ldots, v_{n}\right\}$, where the edge weights correspond to distances between nodes. We target at the common objective to find a permutation of the nodes $\pi$, i.e., a tour, which traverses each node once and returns to the starting one with the shortest distance. On top of the TSP graph, CVRP requires one more node $v_{0}$ as the depot, and prescribes the demands $\left\{\delta_{1}, \delta_{2}, \ldots, \delta_{n}\right\}$ for each node. The objective is to decide routes with the shortest distance for the vehicle(s) satisfying that, 1) each route starts and ends at the depot; 2) each node is traversed by one route; 3) the sum of demands in a route is less than the pre-defined capacity $D$ of vehicle.

Learning-based methods often assume a uniform distribution of nodes to generate the instances, and simply pursue smallest average (optimality) gap of the objective values over them. It inevitably sacrifices the performance on atypical instances that do not follow a majority distribution (i.e., when more than one distribution exist), which may also deteriorate the overall performance.




\subsection{Distributionally Robust Optimization}
Given a parameter family $\Theta$, loss function $\ell$ and training samples $x$ drawn from a distribution $P$, most of existing deep models directly optimize the empirical risk minimization (ERM) as below,
\begin{equation}
\hat{\theta}_{\mathrm{ERM}}:=\underset{\theta \in \Theta}{\arg \min } \mathbb{E}_{x \sim \hat{P}}[\ell(\theta ;x)],
\end{equation}
where $\hat{P}$ is the empirical distribution over the training samples. When the test distribution is same as the one for training, it guarantees that a model trained via ERM performs well for test given sufficient training samples. However, ERM may impair the performance on atypical samples since the model is tuned to emphasize more on the majority samples of the distribution for training. 

Different from ERM, DRO aims at optimizing parameters $\theta$ for a model to achieve more accurate predictions over the test set following diverse or even unknown distributions. It hinges on a more conservative objective that encourages the model to assign higher weights to optimize for atypical samples. Formally, DRO minimizes the \emph{worst} expected loss over an \emph{uncertainty set} of distributions $\mathcal{Q}$ as below,
\begin{equation}
\min _{\theta \in \Theta}\left\{\mathcal{R}(\theta):=\sup _{p \in \mathcal{Q}} \mathbb{E}_{x \sim p}[\ell(\theta; x)]\right\},
\label{eq:dro_1}
\end{equation}
where $\mathcal{Q}$ includes the potential test distributions that we hope the model to perform well on, and $p$ is a possible training distribution sampled form the uncertainty set. By minimizing the \emph{worst-case} loss for all distributions in the uncertainty set $\mathcal{Q}$, we can also expect to achieve desirable overall performance. Note that the objective in Eq.~(\ref{eq:dro_1}) does not necessarily rely on the unknown test distribution $p^\prime$. In the meantime, it is also the upper bound of the test risk \cite{oren2019distributionally}, i.e., $\mathbb{E}_{p^\prime}[\ell(x ; \theta)] \leq \sup _{p \in \mathcal{Q}} \mathbb{E}_{p}[\ell(\theta; x)]$.



\section{Methodology}
In this section, we first exploit the group DRO to train deep models for solving VRPs, which allows them to favorably handle instances of different distributions. Then we fill the gaps for deep models that hinge on reinforcement learning (RL). Finally, we present a convolutional module which could effectively initialize the distribution-aware representations for VRP instances. The overview of our approach is illustrated in Figure~\ref{fig:procedure}.

\subsection{Group DRO for VRPs}


We stipulate that the VRP instances in the training set follow a mixture of distributions rather than a single one, e.g., the majority of typical instances are generated \emph{uniformly} along with the minority of atypical ones sampled from another distribution. While different from those in existing deep models, this setting offers two merits that, 1) it explicitly considers heterogeneous instances, where atypical (yet important) ones may always exist in reality even on a daily basis; 2) the cross-distribution generalization ability could be potentially enhanced. To train deep models with the above setting, DRO is a conceptually appealing option to balance the training on instances of different distributions, since it minimizes the worst expected loss over different distributions in the uncertainty set. It will also potentially empower deep models to tackle the discrepancy between the training and test distributions. However, the effectiveness of DRO often hinges on the choice of the uncertainty set $\mathcal{Q}$, where common practice is to define it as a divergence ball over the entire training distribution. Unfortunately, it may lead to an overly conservative set of potential test distributions that overemphasizes on the minority instances.


To circumvent this issue, we exploit the \emph{group} DRO and explicitly leverage prior knowledge of distributions to group the instances in the training set. In this way, the inferior models that are overwhelmed by unreasonable worst-case distributions due to the divergence ball in DRO, will be avoided~\cite{duchi2019distributionally}. Particularly, we allocate training instances of each distribution into a respective group, termed \emph{distribution group}, and the uncertainty set $\mathcal{Q}$ is defined as the combination of these distribution groups as below, 
\begin{equation}
\mathcal{Q}:=\left\{\sum_{g=1}^{m} q_{g} P_{g}: q \in \Delta_{m}\right\}, 
\label{eq:drogroup}
\end{equation}
where the training set is a mixture of $m$ distribution groups $P_{g}$, indexed by $\mathcal{G}=\{1,2, \ldots, m\}$; $\Delta_{m}$ denotes the $(m-1)$-dimensional probability simplex; $q_g$ denotes the \emph{importance weight} of the $g$-th distribution group.

In light of the above setting, each training instance is associated with a 2-tuple $(x, g)$, where $x$ denotes the input to deep models for an instance and $g$ is its group index. While the group index of each instance in training will enhance the deep models with more desirable cross-distribution generalization ability, this group index is not required during inference. Formally, we update the parametrized model by minimizing the worst empirical expected loss as below,
\begin{equation}
\hat{\theta}_{\mathrm{DRO}}:=\underset{\theta \in \Theta}{\arg \min }\left\{\hat{\mathcal{R}}(\theta):=\max _{g \in \mathcal{G}} \mathbb{E}_{x \sim \hat{P}_{g}}[\ell(\theta ;x)]\right\},
\end{equation}
where $\hat{\mathcal{R}}(\theta)$ refers to the maximum empirical expected loss among all distribution groups, and each group $\hat{P}_{g}$ is an empirical distribution over the corresponding training instances. The above equation could be further reformulated as below,
\begin{equation}
\label{eq:Gdro}
\min _{\theta \in \Theta} \sup _{q \in \Delta_{m}} \sum_{g=1}^{m} q_{g} \mathbb{E}_{(x) \sim P_{g}}[\ell(\theta ;x)],
\end{equation}
where $q$ is a $m$-dimensional trainable vector with the same meaning as the one in Eq.~(\ref{eq:drogroup}). According to Eq.~(\ref{eq:Gdro}), we propose to optimize the parameters of the deep model and importance weights of the distribution groups in an interleaved manner. Specifically, we train a deep model to minimize the loss $\hat{\mathcal{R}}(\theta)$ over distribution groups, while importance weight $q_{g}$ is updated and used to emphasize the expected losses for each distribution group $g$.  





\begin{algorithm}[t]
	\caption{Group DRO for Solving VRPs}
	\label{algo:algorithm}
	\textbf{Input}: Training set $S$, hyperparameter for $M(\theta; x)$\\
	\textbf{Output}: Model parameters $\theta$
	\begin{algorithmic}[1] 
		\STATE Initialize model parameters $\theta^{(0)}$
		\STATE Initialize group weights $q^{(0)}$
		\FOR{t = $1, 2, \dots, T$}
		\STATE Sample a group index $g$ from $(1, \dots, m)$
		\FOR{$t^{\prime} = 1, 2, \dots, T^{\prime} $ }
		\STATE Sample a batch of instances $s$ from group $g$
		\STATE Sample trajectories via $M(\theta^{(t^{\prime})};x)$ solving $s$
		\STATE Calculate the reinforce loss $\ell_r$
		\STATE $\theta^{(t^{\prime})} \leftarrow \theta^{(t^{\prime}-1)}-\eta_{\theta} q_{g}^{(t)} \nabla \ell_r\left(\theta^{(t^{\prime}-1)} ;(x, g)\right)$
		\ENDFOR
		\STATE $q^{\prime} \leftarrow q^{(t-1)} ; q_{g}^{\prime} \leftarrow q_{g}^{\prime} \exp \left(\eta_{q} \ell_r\left(\theta^{(t^{\prime})} ;(x, g)\right)\right)$
		\STATE Renormailize weights by $q^{(t)} \leftarrow q^{\prime} / \sum_{g^{\prime}} q_{g^{\prime}}^{\prime}$
		\ENDFOR
	\end{algorithmic}
\end{algorithm}

\noindent\textbf{Adapt DRO with RL} The DRO with the fashion of supervised learning has been widely studied in the machine learning community~\cite{namkoong2017variance,oren2019distributionally,sagawa2019distributionally}, which could be naturally adapted with the deep models trained in a supervised way for solving VRPs, such as GCN~\cite{joshi2019efficient}. Compared with supervised learning, reinforcement learning (RL) is much preferred given its independence of optimal solution as the ground-truth label. For example, POMO~\cite{kwon2020pomo} trained with reinforcement learning, has achieved state-of-the-art performance among all the deep models. As such, we elaborate on how to adapt the DRO with RL algorithms. 


Without loss of generality, we define a deep model $M$ to sample solutions to VRPs, i.e., $P_\theta(\pi|x)=M(\theta; x)$, where $\theta$ denotes trainable parameters for $M$; $x$ and $\pi$ denote the input and the solution, respectively. We extend REINFORCE~\cite{williams1992simple} to encourage $M$ to generalize well across different distributions. The reinforce loss $\ell_r$ for VRPs is the expected length of the routes in the solution, i.e., $l(\pi)$. Accordingly, we train the deep model parameters $\theta$ using the gradient of the worst expected reinforce loss as below, 



\begin{equation}
\label{eq:A2C_DRO}
\nabla\ell \hspace{-1mm}=\hspace{-1mm}\sup _{q \in \Delta_{m}} \hspace{-1mm}\sum_{g=1}^{m} q_{g} \mathbb{E}_{\substack{M(\theta; x)\\ x\sim P_{g}}}[(l(\pi)-b(x))\nabla logM(\theta; x)],
\end{equation}
where $b(x)$ denotes the baseline to reduce the variance of gradients. We maintain a importance weight $q_g$ for the distribution group $P_g$. The hybrid training with DRO and RL is summarized in Algorithm \ref{algo:algorithm}. Particularly, following the DRO for supervised learning~\cite{sagawa2019distributionally}, we alternately optimize model parameters $\theta$ using stochastic gradient descent (SGD) with the fixed importance weights $q$ (line 9), and update $q$ using exponentiated gradient ascent (line 11). Additionally, we fix $q$ for every $T^{\prime}$ iterations to stabilize the training (lines 5-10). Note that this algorithm can be applied to a wide range of deep RL models for VRPs.


\subsection{Distribution-aware Embedding via CNN}
On the one hand, the nodes in a VRP solution are always connected with the ones in their close proximities, and the local spatial information is critical to reveal the node distribution, which is helpful to enhance the cross-distribution generalization performance for the deep models. On the other hand, while the convolutional neural networks (CNNs) have demonstrated strong capability for learning spatial features in computer vision, directly mapping the graph representation of VRP instances to 2-dimensional images and then applying CNN did not bring obvious benefit~\cite{miki2019solving,ling2020solving}. 

This motivates us to leverage CNN in more elegant ways, i.e., 1) rather than mapping them to 2-dimensional images, we adopt one-dimensional convolution to learn the representation of the nodes; 2) rather than directly relying on the learned representation by CNN, we feed them as the input to the subsequent deep models that are equipped with more advanced architectures (e.g., Transformer in POMO) to learn deeper representations. Accordingly, in our approach, we adopt a convolutional embedding layer to identify the micro-patterns of VRP instances by exploiting the spatial invariance of the nodes. In specific, we embed the input of a node $i$ by performing one-dimensional convolution over its $K$-nearest neighbors in the input graph $G$. 
Firstly,the convolutional layer computes $d_h$-dimensional embedding based on the coordinates of each node $i$ and its $K$-nearest neighbors.  
Then we produce the embedding $h_i$ for node $i$ using a linear projection as
$    h_i = W_1 x_i + W_2 \overline{h_i}, $ 
where $x_i$ is the coordinate of node $i$; $\overline{h_i}$ is the convolutional results of CNN embedding layer; $W_1$ and $W_2$ are trainable matrices. With the embeddings for each node $h_i$ which carries spatial information of its $K$-nearest neighbors, different deep models can process them by the encoder to further produce more informative embeddings for the decoder. 




\section{Experiments and Analysis}

In this section, we conduct experiments to evaluate the efficacy of our approach on the randomly generated instances, and also the ones from two benchmark datasets, i.e., TSPLib and CVRPLib, respectively.



\subsection{Experimental Settings}
Typically, we consider 50 and 100 nodes for TSP and CVRP, respectively. We employ six different types of distributions to generate instances for them, including \emph{uniform}, \emph{explosion}, \emph{implosion}, \emph{expansion}, \emph{cluster}, and \emph{grid}~\cite{bossek2019evolving,zhao2020leveraging}, and also normalize them to the [0,1] square. We generate \emph{uniform} distribution instances as the \emph{typical}  group, and the other one from the five to generate \emph{atypical} instances as the minority group. As such, five combinations of distributions will be initially considered for each problem and size for training, while we will also additionally consider the combination of the six distributions together when applying our approach to solve the complex instances in TSPLib and CVRPLib. Particularly, regarding the \emph{cluster} instances, we set 2 clusters for TSP50 and CVRP50, and 4 for TSP100 and CVRP100, respectively. Regarding CVRP, we compute the demand for each node as $\hat{\delta}_{i}=\delta_{i} / D$, where $\delta_{i}$ is uniformly sampled from $\{1,2, \ldots, 9\}$ and $D$ = 40 and 50 for CVRP50 and CVRP100, respectively. An additional “depot” node without demand is created in a random inside location. 


We apply our approach to two different types of deep models for solving VRPs, i.e., GCN~\cite{joshi2019efficient} \footnote{https://github.com/chaitjo/graph-convnet-tsp} and POMO~\cite{kwon2020pomo} \footnote{https://github.com/yd-kwon/POMO}, which are trained in the fashion of supervised learning and reinforcement learning, respectively. We term GCN and POMO that are equipped with our group DRO and CNN as \emph{DROG} and \emph{DROP}, respectively. Pertaining to \emph{DROG}, we follow the suggestion stated in the original work of GCN by modifying the output to predict the next node to visit rather than an adjacency matrix. Thus, we train it as an auto-regressive model while still in the supervised way. The training labels (distance of the route) of TSP and CVRP are obtained by Concorder~\cite{applegate2006concorde} and LKH3~\cite{helsgaun2000effective} respectively. We also set the inner training loop number $T^{\prime}$ in Algorithm~\ref{algo:algorithm}  to $1$, and use the cross-entropy loss to optimize edge probabilities from output layer for the sake of GCN.
Pertaining to \emph{DROP}, the original POMO introduces a data augmentation technique to exploit the symmetries of VRP instances in inference, where we adopt $\times 8$ augmentation following its original setting. We also apply group adjustments with strong $\ell_2$ penalties in group DRO as did in \cite{sagawa2019distributionally}. Regarding the convolutional embedding layer by CNN, the input length for each node equals to 2 (coordinate) for TSP and 3 (coordinate+demand) for CVRP. We extract the spatial pattern of $K$=10 nearest nodes with the kernel size 11 and set the number of kernels to 128. The output dimension is fixed to 128.



Our approach is implemented in PyTorch 1.2 \cite{NEURIPS2019_9015} with Python 3.7. We run the experiments on the device with a single Nvidia GeForce RTX 2080Ti GPU and a single CPU of an Intel Xeon i9-10940X CPU at 3.3 GHz. We use the SGD optimizer with minibatches and a momentum. The learning rate $\eta$ is $10^{-4}$ for all experiments. Training time varies with the size of the problem, from a couple of hours to a week. Taking TSP100 as an example, one training epoch consumes about 11 minutes. We trained the models up to 2,000 epochs ($\sim 10$ days) to observe full convergence, although most of them are converged within 300 epochs ($\sim 2$ day). However, the inference time is much shorter, which is almost the same as the respective original models, as shown in the following tables. Note that all baseline deep models including GCN and POMO (and AM) are re-trained in our device, considering the instances used in this paper are essentially different from the ones in their original works.




\subsection{Comparison with Baselines}

\begin{table*}[hbt!]  \small
\setlength{\tabcolsep}{1.4pt}
	\centering
	\caption{Results for Comparison with Baselines}	
	\begin{threeparttable}
	    \scalebox{0.94}{
		\begin{tabular}{@{}p{0.1cm}c|ccccc|ccccc||ccccc|ccccc@{}}
			\toprule
			\multicolumn{2}{c|}{Distribution}& \multicolumn{5}{c|}{TSP50} & \multicolumn{5}{c||}{TSP100}   & \multicolumn{5}{c|}{CVRP50} & \multicolumn{5}{c}{CVRP100} \\
			\multicolumn{2}{c|}{$\&$Method}& Obj. & Gap & Obj$^\star$ & Gap$^\star$ & Time  & Obj. & Gap & Obj$^\star$ & Gap$^\star$ & Time & Obj. & Gap & Obj$^\star$ & Gap$^\star$ & Time & Obj. & Gap & Obj$^\star$ & Gap$^\star$ & Time\\ \midrule\midrule
\multirow{6}{*}{\rotatebox[origin=c]{90}{\textbf{Explosion}}} & Solvers & 6.34          & 0.00\%          & 6.69          & 0.00\%          & 15m & 8.58          & 0.00\%          & 9.15          & 0.00\%          & 1h  & 12.02          & 0.00\%          & 12.30          & 0.00\%          & 5h  & 17.18          & 0.00\%          & 17.53          & 0.00\%          & 11h \\
          & AM      & 6.60          & 4.10\%          & 7.11          & 6.28\%          & 8m  & 9.01          & 5.01\%          & 9.96          & 8.85\%          & 31m & 12.31          & 2.41\%          & 12.83          & 4.31\%          & 14m & 17.79          & 3.55\%          & 18.34          & 4.62\%          & 52m \\
          & GCN     & 6.47          & 2.05\%          & 6.94          & 3.74\%          & 25m & 8.87          & 3.38\%          & 9.70          & 6.01\%          & 52m & -              & -               & -              & -               & -   & -              & -               & -              & -               & -   \\
          & DROG    & 6.43          & 1.42\%          & 6.75          & 0.90\%          & 33m & 8.74          & 1.86\%          & 9.46          & 3.39\%          & 1h  & 12.17          & 1.25\%          & 12.50          & 1.63\%          & 38m & \textbf{17.55} & \textbf{2.15\%} & 17.96          & 2.45\%          & 2h  \\
          & POMO    & 6.52          & 2.84\%          & 7.05          & 5.38\%          & 19s & 8.84          & 3.03\%          & 9.63          & 5.25\%          & 1m  & 12.26          & 2.00\%          & 12.69          & 3.17\%          & 32s & 17.64          & 2.68\%          & 18.14          & 3.48\%          & 3m  \\
          & DROP    & \textbf{6.38} & \textbf{0.63\%} & \textbf{6.71} & \textbf{0.30\%} & 18s & \textbf{8.73} & \textbf{1.75\%} & \textbf{9.40} & \textbf{2.73\%} & 1m  & \textbf{12.15} & \textbf{1.08\%} & \textbf{12.46} & \textbf{1.30\%} & 31s & \textbf{17.55} & \textbf{2.15\%} & \textbf{17.95} & \textbf{2.40\%} & 3m  \\

			\midrule
			\multirow{6}{*}{\rotatebox[origin=c]{90}{\textbf{Implosion}}} & Solvers & 6.40          & 0.00\%          & 6.74          & 0.00\%          & 15m & 8.76          & 0.00\%          & 9.50          & 0.00\%          & 1h  & 12.11          & 0.00\%          & 12.45          & 0.00\%          & 5h  & 17.25          & 0.00\%          & 17.61          & 0.00\%          & 10h \\
          & AM      & 6.64          & 3.75\%          & 7.23          & 7.27\%          & 8m  & 9.12          & 4.11\%          & 10.21         & 7.47\%          & 33m & 12.56          & 3.72\%          & 12.93          & 3.86\%          & 13m & 17.70          & 2.61\%          & 18.42          & 4.60\%          & 52m \\
          & GCN     & 6.50          & 1.56\%          & 7.13          & 5.79\%          & 25m & 8.94          & 2.05\%          & 9.93          & 4.53\%          & 58m & -              & -               & -              & -               & -   & -              & -               & -              & -               & -   \\
          & DROG    & 6.45          & 0.78\%          & 6.79          & 0.74\%          & 27m & 8.83          & 0.80\%          & 9.64          & 1.47\%          & 1h  & \textbf{12.26} & \textbf{1.24\%} & \textbf{12.55} & \textbf{0.80\%} & 30m & 17.58          & 1.91\%          & 18.06          & 2.56\%          & 3h  \\
          & POMO    & 6.57          & 2.66\%          & 7.16          & 6.23\%          & 22s & 8.89          & 1.48\%          & 9.80          & 3.16\%          & 2m  & 12.43          & 2.64\%          & 12.71          & 2.09\%          & 30s & 17.65          & 2.32\%          & 18.25          & 3.63\%          & 3m  \\
          & DROP    & \textbf{6.43} & \textbf{0.47\%} & \textbf{6.78} & \textbf{0.59\%} & 22s & \textbf{8.81} & \textbf{0.57\%} & \textbf{9.58} & \textbf{0.84\%} & 2m  & 12.28          & 1.40\%          & 12.60          & 1.20\%          & 29s & \textbf{17.52} & \textbf{1.57\%} & \textbf{18.02} & \textbf{2.33\%} & 3m  \\

			\midrule
			\multirow{6}{*}{\rotatebox[origin=c]{90}{\textbf{Expansion}}} & Solvers & 6.13          & 0.00\%          & 6.19          & 0.00\%          & 13m & 8.19          & 0.00\%          & 9.28          & 0.00\%          & 1h  & 11.86          & 0.00\%          & 12.14          & 0.00\%          & 5h  & 16.98          & 0.00\%          & 17.37          & 0.00\%          & 9h  \\
          & AM      & 6.36          & 3.75\%          & 6.55          & 5.82\%          & 7m  & 8.46          & 3.30\%          & 9.72          & 4.74\%          & 28m & 12.23          & 3.12\%          & 12.69          & 4.53\%          & 12m & 17.66          & 4.00\%          & 18.23          & 4.95\%          & 44m \\
          & GCN     & 6.24          & 1.79\%          & 6.53          & 5.49\%          & 23m & 8.41          & 2.69\%          & 9.53          & 2.69\%          & 54m & -              & -               & -              & -               & -   & -              & -               & -              & -               & -   \\
          & DROG    & \textbf{6.18} & \textbf{0.82\%} & 6.26          & 1.13\%          & 24m & \textbf{8.34} & \textbf{1.83\%} & \textbf{9.36} & \textbf{0.86\%} & 1h  & \textbf{11.95} & \textbf{0.76\%} & \textbf{12.24} & \textbf{0.82\%} & 28m & 17.32          & 2.00\%          & 17.75          & 2.19\%          & 2h  \\
          & POMO    & 6.29          & 2.61\%          & 6.61          & 6.79\%          & 19s & 8.40          & 2.56\%          & 9.50          & 2.37\%          & 1m  & 12.08          & 1.85\%          & 12.49          & 2.88\%          & 26s & 17.55          & 3.36\%          & 18.04          & 3.86\%          & 2m  \\
          & DROP    & 6.20          & 1.14\%          & \textbf{6.23} & \textbf{0.65\%} & 19s & \textbf{8.34} & \textbf{1.83\%} & 9.35          & 0.75\%          & 1m  & 11.97          & 0.93\%          & 12.28          & 1.15\%          & 25s & \textbf{17.30} & \textbf{1.88\%} & \textbf{17.73} & \textbf{2.07\%} & 2m  \\

			\midrule
			\multirow{6}{*}{\rotatebox[origin=c]{90}{\textbf{Cluster}}} & Solvers & 5.88          & 0.00\%          & 6.17          & 0.00\%          & 10m & 7.44          & 0.00\%          & 7.61          & 0.00\%          & 1h  & 11.68          & 0.00\%          & 11.83          & 0.00\%          & 5h  & 16.69          & 0.00\%          & 16.98          & 0.00\%          & 8h  \\
          & AM      & 6.12          & 4.08\%          & 6.67          & 8.10\%          & 6m  & 7.81          & 4.97\%          & 7.97          & 4.73\%          & 26m & 12.04          & 3.08\%          & 12.47          & 5.41\%          & 10m & 17.30          & 3.65\%          & 17.97          & 5.83\%          & 40m \\
          & GCN     & 6.02          & 2.38\%          & 6.34          & 2.76\%          & 22m & 7.69          & 3.36\%          & 7.92          & 4.07\%          & 50m & -              & -               & -              & -               & -   & -              & -               & -              & -               & -   \\
          & DROG    & 5.91          & 0.51\%          & 6.24          & 1.13\%          & 23m & 7.64          & 2.69\%          & 7.77          & 2.10\%          & 1h  & 11.84          & 1.37\%          & 12.10          & 2.28\%          & 27m & 17.09          & 2.40\%          & 17.50          & 3.06\%          & 2h  \\
          & POMO    & 6.07          & 3.23\%          & 6.33          & 2.59\%          & 18s & 7.68          & 3.23\%          & 7.80          & 2.50\%          & 1m  & 11.88          & 1.71\%          & 12.32          & 4.14\%          & 23s & 17.22          & 3.18\%          & 17.73          & 4.42\%          & 2m  \\
          & DROP    & \textbf{5.90} & \textbf{0.34\%} & \textbf{6.22} & \textbf{0.81\%} & 18s & \textbf{7.58} & \textbf{1.88\%} & \textbf{7.73} & \textbf{1.58\%} & 1m  & \textbf{11.79} & \textbf{0.94\%} & \textbf{12.03} & \textbf{1.69\%} & 23s & \textbf{17.01} & \textbf{1.92\%} & \textbf{17.42} & \textbf{2.59\%} & 2m  \\

			\midrule
			\multirow{6}{*}{\rotatebox[origin=c]{90}{\textbf{Grid}}}      & Solvers & 6.02          & 0.00\%          & 6.28          & 0.00\%          & 11m & 7.85          & 0.00\%          & 7.99          & 0.00\%          & 1h  & 11.72          & 0.00\%          & 11.97          & 0.00\%          & 5h  & 16.71          & 0.00\%          & 17.05          & 0.00\%          & 9h  \\
          & AM      & 6.24          & 3.65\%          & 6.65          & 5.89\%          & 7m  & 8.13          & 3.57\%          & 8.51          & 6.51\%          & 28m & 12.14          & 3.58\%          & 12.52          & 4.59\%          & 10m & 17.32          & 3.65\%          & 17.90          & 4.99\%          & 43m \\
          & GCN     & 6.19          & 2.82\%          & 6.38          & 1.59\%          & 22m & 8.09          & 3.06\%          & 8.40          & 5.13\%          & 51m & -              & -               & -              & -               & -   & -              & -               & -              & -               & -   \\
          & DROG    & \textbf{6.06} & \textbf{0.66\%} & \textbf{6.30} & \textbf{0.32\%} & 23m & 7.96          & 1.40\%          & 8.15          & 2.00\%          & 1h  & 11.83          & 0.94\%          & 12.12          & 1.25\%          & 27m & 17.08          & 2.21\%          & 17.51          & 2.70\%          & 2h  \\
          & POMO    & 6.20          & 2.99\%          & 6.41          & 2.07\%          & 18s & 8.01          & 2.04\%          & 8.32          & 4.13\%          & 1m  & 11.92          & 1.71\%          & 12.35          & 3.17\%          & 25s & 17.21          & 2.99\%          & 17.75          & 4.11\%          & 2m  \\
          & DROP    & 6.10          & 1.33\%          & 6.33          & 0.80\%          & 17s & \textbf{7.93} & \textbf{1.02\%} & \textbf{8.09} & \textbf{1.25\%} & 1m  & \textbf{11.80} & \textbf{0.68\%} & \textbf{12.08} & \textbf{0.92\%} & 25s & \textbf{17.00} & \textbf{1.74\%} & \textbf{17.47} & \textbf{2.46\%} & 2m  \\ 
            \bottomrule
		\end{tabular}}
		\begin{tablenotes} \small
			\item[1] The gap is computed based on Concorder for TSP and LKH3 for CVRP respectively.
			\item[2] \textbf{$\star$} means the average results over atypical instances; \textbf{Bold} means the best result from deep models.			
		\end{tablenotes}		
	\end{threeparttable}
	\label{tb:large}
\end{table*}

\begin{table*}[ht]  \small
\setlength{\tabcolsep}{1.4pt}
\renewcommand{\arraystretch}{1.15}
	\centering
	\caption{Results for Ablation Study}	
	\begin{threeparttable}
		\begin{tabular}{@{}l|cccc|cccc|cccc|cccc|cccc@{}}
			\toprule
			& \multicolumn{4}{c|}{Concorde
} & \multicolumn{4}{c|}{POMO}   & \multicolumn{4}{c|}{POMO+CNN} & \multicolumn{4}{c|}{POMO+DRO} & \multicolumn{4}{c}{DROP}   \\
			Distribution & Obj. & Gap & Obj$^\star$ & Gap$^\star$  & Obj. & Gap & Obj$^\star$ & Gap$^\star$& Obj. & Gap & Obj$^\star$ & Gap$^\star$ & Obj. & Gap & Obj$^\star$ & Gap$^\star$ & Obj. & Gap & Obj$^\star$ & Gap$^\star$\\ \midrule\midrule
Explosion & 8.58 & 0.00\% & 9.15 & 0.00\% & 8.84 & 3.03\% & 9.63 & 5.25\% & 8.80 & 2.56\% & 9.62 & 5.14\% & 8.76 & 2.10\% & 9.45 & 3.28\% & \textbf{8.73} & \textbf{1.75\%} & \textbf{9.40} & \textbf{2.73\%} \\
Implosion & 8.76 & 0.00\% & 9.50 & 0.00\% & 8.89 & 1.48\% & 9.80 & 3.16\% & 8.83 & 0.80\% & 9.76 & 2.74\% & 8.83 & 0.80\% & 9.59 & 0.95\% & \textbf{8.81} & \textbf{0.57\%} & \textbf{9.58} & \textbf{0.84\%} \\
Expansion & 8.19 & 0.00\% & 9.28 & 0.00\% & 8.40 & 2.56\% & 9.50 & 2.37\% & 8.37 & 2.20\% & 9.45 & 1.83\% & 8.35 & 1.95\% & 9.37 & 0.97\% & \textbf{8.34} & \textbf{1.83\%} & \textbf{9.35} & \textbf{0.75\%} \\
Cluster   & 7.44 & 0.00\% & 7.61 & 0.00\% & 7.68 & 3.23\% & 7.80 & 2.50\% & 7.61 & 2.28\% & 7.79 & 2.37\% & 7.62 & 2.42\% & 7.75 & 1.84\% & \textbf{7.58} & \textbf{1.88\%} & \textbf{7.73} & \textbf{1.58\%} \\
Grid      & 7.85 & 0.00\% & 7.99 & 0.00\% & 8.01 & 2.04\% & 8.32 & 4.13\% & 7.97 & 1.53\% & 8.27 & 3.50\% & 7.95 & 1.27\% & 8.12 & 1.63\% & \textbf{7.93} & \textbf{1.02\%} & \textbf{8.09} & \textbf{1.25\%} \\
		
            \bottomrule
		\end{tabular}
	\end{threeparttable}
	\label{tb:ablation}
\end{table*}


\begin{table}[htb!] \small
\setlength{\tabcolsep}{5.4pt}
\renewcommand{\arraystretch}{1}
\centering 
\caption{Generalization Results on TSPLib and CVRPLib} 
\begin{threeparttable}
\begin{tabular}{l|ccc|cccc}
\toprule 
\textbf{Instance} & Opt. & AM &
POMO  & DROG & DROP \\ 
\midrule
Eil51 & 426 & 439 & 436 & 427  & \textbf{426}\\
Berlin52  & 7542  & 8352  & 7836  & 7553  & \textbf{7544} \\
St70  & 675   & 691   & 683   & 684   & \textbf{679}\\
rat99  & 1211  & 1340  & 1286  & \textbf{1233}  & 1248\\

KroA100 &21282 & 46621& 38452& 25196 &\textbf{24623}\\
KroB100 &22141  &37921 &33521 &26583 &\textbf{24874}\\
KroC100 & 20749 &  34258 &30736 &\textbf{24343}& 24785\\
KroD100 & 21294 & 36141 & 29512 & 23633 & \textbf{23257}\\
KroE100  &22068 & 29628& 26829 & 26289& \textbf{26057} \\

rd100  & 7910  & 8252  & 8180  & 8137  & \textbf{8043}\\
lin105 & 14379 & 15148 & 14922 & 14876 & \textbf{14688}\\
pr107 & 44303 & 53846 & 52846 & \textbf{46572} & 47853 \\
ch150  & 6528  & 6930  & 6844  & 6792  & \textbf{6709}\\ 
rat195  & 2323  & 2612  & 2554  & 2466  & \textbf{2403}\\ 
kroA200 & 29368 & 35637 & 34972 & 34682 & \textbf{34275}\\
\midrule\midrule
X-n101-k25 & 27591 & 38264 & 29484 & 29167 & \textbf{28949}\\ 

X-n106-k14  & 26362 & 27923 & 27762 & 27331 & \textbf{27308} \\

X-n110-k13 & 14971 & 16320 & 15896 & \textbf{15226} & 15386\\ 
X-n115-k10 & 12747 & 14055 & 13952 & 13921 & \textbf{13783}\\ 
X-n120-k6 & 13332 & 14456 & 14351 & 14227 & \textbf{14058}\\ 

X-n125-k30 & 55539 & 74329 & 69560 & 64596 & \textbf{61382} \\

X-n129-k18 & 28940 & 30869 & 30155 & 30181 & \textbf{30075}\\ 

X-n134-k13 & 10916 & 13952 & 13483 & 13117 & \textbf{12846} \\

X-n139-k10 & 13590 & 14893 & 14132 & 14153 & \textbf{13979}\\ 

X-n143-k7 & 15700 & 18251 & 17923 & \textbf{17547} & 17682\\

X-n153-k22 & 21220 & 38423 & 26386 & 24591 & \textbf{24386}\\
X-n157-k13 & 16876 & 22051 & 19978 & 18993 & \textbf{18378}\\ 
X-n181-k23 &25569 & 27826 & 27428 & 27232 & \textbf{27094} \\
X-n190-k8 & 16980 & 37820 & 22310 & 20682 & \textbf{19864}\\ 
X-n200-k36 &58578 & 76528 & 73135 & 68283 & \textbf{64921} \\
\midrule
\midrule 
Avg. Gap & 0.00\% & 29.93\% & 17.57\% & 9.68\% & \textbf{8.08\%}\\ 
\bottomrule 
\end{tabular}
\begin{tablenotes} \small
	\item[1] \textbf{Bold} means the best result from deep models.
\end{tablenotes}
\end{threeparttable}
\label{tb:cvrplib} 
\end{table}

We compare DROG and DROP with a variety of baselines, 1) Concorde, a specialized exact solver for TSP; 2) LKH3, a strong heuristic solver with state-of-the-art performance on many VRP variants; 3) attention model (AM)~\cite{kool2018attention}, the Transformer based deep model which achieves desirable results on both TSP and CVRP and recognized as a milestone; 4) GCN (original), a graph convolutional network for VRP which is trained in a supervised way and outputs the probabilities of edges that will appear in the optimal route(s) in the format of adjacency matrix; 5) POMO (original), a recent deep model developed based on AM, which is trained by RL and achieves state-of-the-art results among the deep models. Following the common settings in the works of the deep models, we use Concorde to solve TSP instances, and LKH3 to solve CVRP instances. 

We train both DROG and DROP with five pairs of distribution groups in DRO. Specifically, the training set for each pair comprises 100,000 typical instances from the \emph{uniform} distribution and 10,000 atypical instances from one of the distributions including \emph{explosion}, \emph{implosion}, \emph{expansion}, \emph{cluster}, and \emph{grid}, respectively. 
The test set comprises 10,000 typical instances and 1,000 atypical instances sampled from the same pair of distribution groups used in training. Note that, the label or index of the distribution group is not required during inference. All the baseline deep models are trained and tested using the same instances as ours. Since GCN did not solve CVRP in its original work, we do not apply it to solve CVRP either. 

We record the performance in terms of objective value, (optimality) gap and computation time, which are averaged over the instances in each test set. Particularly, we also record the objective value and gap for the atypical instances to indicate the worst-case performance. All results are summarized in Table~\ref{tb:large}. We see that our models (DROG and DROP) significantly outperform all other learning-based baselines on the five pairs of distributions, achieving much smaller objective values and gaps on both TSP and CVRP of different sizes. The superiority is more obvious for the atypical instances from the minority group, where one motivating observation is that the gaps delivered by DROP are much smaller than those by POMO, e.g., 1.25\% vs 4.13\% (\emph{grid}) and 2.59\% vs 4.42\% (\emph{cluster}) on TSP100 and CVRP100, respectively. We also found that while the original GCN is inferior to POMO, DROG significantly surpasses POMO for all problems and achieves even better results than DROP occasionally, e.g., TSP50 (\emph{grid}), TSP100 (\emph{expansion}) and CVRP50 (\emph{implosion}).


Although the deep models like AM, GCN and POMO have demonstrated desirable performance when they are trained and tested solely using the \emph{uniform} distribution (as reported in their original works), their overall and worst-case performance obviously deteriorate in the presence of atypical instances (as shown in Table~\ref{tb:large}). One of the major reasons is that the models would be overwhelmed by the typical instances (majority) even if they also consider the atypical instances (minority) in training. In contrast, our approach elegantly balances the training on different distribution groups (i.e. either the majority or minority one) by leveraging the group DRO, and enhance the distribution-aware embedding by CNN. 
Moreover, our approach is versatile to different deep models either trained by supervised or reinforcement learning, such as GCN and POMO used in our experiments. Besides, our approach would not incur much additional time for inference, as suggested in Table~\ref{tb:large}.


\subsection{Ablation Study}

To further verify the effectiveness of the components in our approach, i.e., group DRO and CNN, we conduct an ablation study on TSP100 by separately removing the corresponding component from our DROP. The experimental setting is similar to the previous one, and the results are displayed in Table~\ref{tb:ablation}. When POMO is solely equipped with CNN, we observe that the average objective values for the overall instances and the atypical instances are both decreased. Benefiting from the initial distribution-aware embedding, it indicates that the convolutional embedding layer can help learn better spatial patterns of nodes and improve the generalization performance. On the other hand, when POMO is solely equipped with group DRO, it can also ameliorate the performance over the original POMO on both the overall instances and the atypical instances. We also found that the advantage brought by DRO (POMO+DRO) is more significant than that of CNN (POMO+CNN), especially on the atypical instances, since it is specialized to improve the performance for instances of the worst-case group. Finally, when equipped with both CNN and group DRO, our DROP achieves further better results than the respective variants and the original model, which well justified the effectiveness of the two components that could jointly promote the performance on both the overall and atypical instances.


\subsection{Evaluation on Benchmark Dataset}

We continue to evaluate DROG and DROP on the public benchmark dataset, i.e., TSPLib~\cite{reinelt1991tsplib} and CVRPLib~\cite{uchoa2017new}, whose instances may follow various distributions that are totally different from ours. We train our models on 150,000 instances, which include 100,000 instances from \emph{uniform} distribution, and 50,000 instances from the other five, with 10,000 for each (although they do not need to be the same). Then we apply our models trained on TSP100 and CVRP100 to solve some representative instances, whose sizes range from 51 to 200. We separately display results for TSP and CVRP in the upper and lower half of Table~\ref{tb:cvrplib}. The baseline models are also re-trained using the same instances. It is revealed that DROG and DROP achieve near-optimal solutions on both TSP and CVRP. Both DROG and DROP, with average gaps of 9.68\% and 8.08\%, produce much better results than that of AM and POMO, with average gaps of 29.93\% and 17.57\%, respectively, although they are using the same training set as ours. This superiority suggests that, our approach equipped with group DRO and CNN, allows the deep models to favorably generalize to different distributions and sizes. Last but not least, the inference of DROG and DROP are almost as efficient as AM and POMO, respectively, indicating that our approach introduces negligible additional computation overhead.

\section{Conclusion}
In this paper, we exploit group DRO (distributionally robust optimization) to enhance the cross-distribution generalization ability for deep models that are used to solve VRPs. We also leverage an elegant CNN to learn the initial distribution-aware representations of VRP instances. Our approach is readily applicable with either supervised or reinforcement learning and evaluated with two well-known deep models, i.e., GCN and POMO. Empirical results show that our approach significantly improves the cross-distribution generalization performance and outstrips other learning based methods on both the synthesized and benchmark dataset. The ablation study also verifies the efficacy of components in our approach. In future, we will investigate more flexible combinations of distributions and tackle problems of larger scales.  

\section{Acknowledgments}

This study is supported under the RIE2020 Industry Alignment Fund – Industry Collaboration Projects (IAF-ICP) Funding Initiative, as well as cash and in-kind contribution from Singapore Telecommunications Limited (Singtel), through Singtel Cognitive and Artificial Intelligence Lab for Enterprises (SCALE@NTU). It is also partially supported by the National Natural Science Foundation of China under Grant 61803104.

\bibliography{reference}

\end{document}